\documentclass[lettersize,journal]{IEEEtran}

\usepackage{amsmath,amsfonts}
\usepackage{algorithm}
\usepackage{algpseudocode}

\usepackage[utf8]{inputenc}
\usepackage[T1]{fontenc}

\usepackage[dvipsnames]{xcolor}
\definecolor{mc1}{HTML}{000000}  

\usepackage{siunitx}

\usepackage{booktabs}
\usepackage{multirow}
\usepackage{array}
\usepackage{colortbl}
\usepackage{float}
\usepackage{placeins}
\usepackage{textcomp}
\usepackage{verbatim}
\usepackage{pifont}

\usepackage{graphicx}
\usepackage{subcaption}  
\usepackage{stfloats}
\usepackage{cuted}
\usepackage{mwe}  
\usepackage{lipsum}  

\usepackage[hyphens]{url}
\usepackage{hyperref}
\usepackage[noadjust]{cite}
\usepackage{comment}

\usepackage[font=footnotesize, labelsep=period]{caption}
\usepackage[Conny]{fncychap}

\begin{document}

\captionsetup[figure]{labelfont=normalfont, labelformat=simple, labelsep=period, name=Fig.}

\title{Constrained Optimal Fuel Consumption of HEVs under Observational Noise}

\author{%
Shuchang Yan\textsuperscript{†} and Haoran Sun\textsuperscript{†}%
\thanks{\textsuperscript{†}The authors contributed equally to this work.}%
\thanks{Shuchang Yan is with the AI Technology Research and Development Department, 
Automotive New Technology Research Institute, BYD Auto Industry Company Limited, 
Shenzhen 518118, China. 
Haoran Sun is with Tsinghua University, Beijing 100084, China.}%
\thanks{This work was supported by the National Key Research and Development Program of 
China under Grant No. 2021ZD0114100.}%
}

\markboth{Journal of \LaTeX\ Class Files,~Vol.~1, No.~1, August~2025}%
{Shell \MakeLowercase{\textit{et al.}}: A Sample Article Using IEEEtran.cls for IEEE Journals}

\maketitle

\begin{abstract}
In our prior work, we evaluated the minimum fuel consumption of a hybrid electric vehicle (HEV) under a battery state of charge (SOC) balance constraint, assuming precise SOC observations and accurate reference speed profiles. The constrained optimal fuel consumption (COFC) problem was solved using a constrained reinforcement learning (CRL) approach. However, SOC measurements are often noisy or inaccurate due to sensor errors, and reference speeds may deviate from actual values. Under these conditions, we aim to minimize fuel consumption while maintaining SOC balance subject to observational noise in SOC and reference speed. We address this challenge by reformulating the COFC problem to incorporate observational noise, modeled by a uniform distribution, and applying robust CRL methods with a generalized and structured training framework to handle it. We validate the approach through simulations on the Toyota Prius hybrid system (THS) under both the New European Driving Cycle (NEDC) and the Worldwide Harmonized Light Vehicles Test Cycle (WLTC). Fuel consumption and SOC constraint satisfaction are evaluated under varying noise conditions, showing stable training performance and revealing that both SOC and speed noise can affect overall fuel consumption to varying degrees. To our knowledge, this is the first work to formally address a widely encountered issue in HEV evaluation and operation: how observational noise, \textbf{prevalent} in dynamometer testing and predictive energy control (PEC) scenarios, affects constrained optimal fuel consumption.
\end{abstract}


\color{black}
\section*{Nomenclature}
\addcontentsline{toc}{section}{Nomenclature}
\begin{IEEEdescription}[\IEEEusemathlabelsep\IEEEsetlabelwidth{$V_1,V_2,V_3$}]
	
    \item[\textcolor{mc1}{ON}] \textcolor{mc1}{Observational noise.}
    
    \item[OEM] Original equipment manufacturer.
    \item[OCV] Open-circuit voltage.
    \item[$\pi_{\theta}$] Policy with parameter $\theta$.
    \item[$\Pi$] Policy set.
    \item[$\mathcal{S}$] State set.
    \item[\(\circ\)] Function composition.

  \item[{$\{ \}_N$}] Batch of $N$ elements within the bracket.
  \item[$\mathcal{U}(x, y)$] Uniform distribution with a range from $x$ to $y$.

  \item[PID] Proportional–integral–derivative.
  \item[TD]  Temporal difference.


\item[$T$] Rollout.
\item[$M$] Total number of policy optimization steps.
\item[$N_{train}$] Total number of training epochs.
\item[$\ell_{\text{ppol}}$] Loss function for the Lagrangian-based PPO.
\item[$\nu(s)$] Noise function on $s$.

\end{IEEEdescription}
\color{black}

\section{Introduction}

The COFC problem focuses on controlling HEVs to minimize fuel consumption while maintaining battery SOC balance. This issue, commonly encountered in the automotive industry, is first formulated and solved as a constrained optimal control problem from the CRL perspective in our seminal work~\cite{yancofc}. While the previous formulation assumes accurate observations, real-world signals such as SOC and reference speed are often corrupted by observational noise. In this work, we relax the assumption of perfect observation and examine how such noise affects optimal fuel consumption decisions. This raises a practical yet fundamental question:

\vskip 0.5 cm
\begin{center}
\parbox{0.80\linewidth}{
\centering
\color{black}\itshape
When observations such as SOC and reference speed are affected by noise, how should the COFC problem be described and solved?
}
\end{center}
\vskip 0.3 cm

We refer to this question as the Constrained Optimal Fuel Consumption under Observational Noise (COFC-ON) problem. Here, ``ON'' denotes observational noise, which arises from sensor inaccuracies and estimation errors, both frequently encountered in HEV evaluation and operation. These uncertainties stem from limitations in measurement and forecasting. For example, SOC cannot be measured directly and is typically estimated from terminal voltage, current, and internal resistance, all of which are influenced by temperature, operating conditions, and sensor precision, making SOC estimation inherently nontrivial. The reference speed also plays a critical role in two representative engineering scenarios. First, during the transition from trial production to mass production, engineers need to conduct chassis dynamometer tests according to regulatory standards, where the driver follows a real-time speed profile shown on a screen. These standards include China’s GB/T 19754-2005~\cite{GBT19754-2005}, SAE J1711~\cite{SAEJ1711}, and Japan’s JC08~\cite{JC08_Test_Cycle}. Second, Predictive Energy Control (PEC), now widely adopted in industry, employs Model Predictive Control based on forecasted speed profiles obtained from commercial map services such as Gaode~\cite{amap_mobile} or Baidu~\cite{baidu_map_location}, which may deviate slightly from actual driving conditions. In both cases, SOC and reference speed are subject to observational noise. Understanding how such deviations affect fuel optimal decisions remains an open problem that has yet to be explicitly formulated or properly addressed in prior vehicle energy control research.

Given observational noise in battery SOC and reference speed, it is important to evaluate its impact on optimal fuel consumption. This consideration motivates the use of learning-based methods for vehicle energy control, with attention to RL, CRL, and their practical implementation aspects.

While RL has made notable progress in recent years~\cite{mnih2013playing, silver2017mastering}, its deployment in real-world applications remains challenging, particularly in handling state constraints. In the vehicle domain, \cite{lian2020rule} incorporates expert knowledge into the Deep Deterministic Policy Gradient (DDPG) method to improve energy control. \cite{zhou2021knowledge} applies DDPG to hybrid powertrain optimization and real-time control for plug-in hybrid electric vehicles (PHEVs). \cite{liu2023safe} uses the distributed PPO method to address safety-constrained energy control under standard driving cycles. \cite{liu2024deep} adopts the Soft Actor-Critic (SAC) method to enhance energy management for heavy-duty HEVs in hybrid action spaces. \cite{shuai2024optimal} presents an ensemble learning-based method to balance fuel consumption and SOC constraints under various driving conditions. In these works, SOC control is included as a penalty term in the RL objective. While RL is effective in optimizing fuel consumption, it cannot strictly enforce key state constraints, such as maintaining SOC within the desired range, thus limiting its reliability in practical applications.

To overcome these limitations, constrained reinforcement learning (CRL) has been developed to jointly optimize reward and constraint performance~\cite{liu2020policy, chen2021instrumented}. Common approaches include the Lagrangian method, which solves the problem using a primal-dual formulation~\cite{chow2017risk, chow2018lyapunov, stooke2020responsive}, and constrained variational policy optimization (CVPO)~\cite{liu2022constrained}, which enforces feasibility through the Expectation-Maximization (EM) framework. In our previous work~\cite{yancofc}, CRL was applied to HEV energy control, addressing two practical objectives: (i) minimizing fuel consumption with SOC balancing under prescribed speed profiles, and (ii) minimizing fuel consumption under the PEC function. While this framework effectively handles fuel optimization and SOC balance under fixed reference speed profiles and accurate SOC observations, it does not account for observational noise typically observed in both roller dynamometer testing and PEC scenarios. This motivates an extension of the CRL formulation to incorporate such observational noise.

\color{black}

Considering the above, we survey robust RL and CRL approaches. Robustness refers to controlling the plant model under unknown dynamics, disturbances, or both~\cite{chandrasekharan1996robust}. Recent robust RL works address action perturbation~\cite{tessler2019action}, reward perturbation~\cite{wang2020reinforcement, lin2020model, eysenbach2021maximum}, plant dynamics uncertainty~\cite{wang2020reinforcement, huang2022robust}, and observational perturbation~\cite{zhang2020robust, zhang2021robust, liang2022efficient, pattanaik2017robust}—which directly relates to our problem—affected by inaccuracies in observed battery SOC and reference velocity.
\cite{zhang2020robust} introduces MAD perturbation on states, finding perturbed states by maximizing the Kullback--Leibler divergence between original and perturbed states. \cite{zhang2021robust} extends this by training an auxiliary agent to implement perturbations alongside the primary agent. \cite{liang2022efficient} estimates and optimizes the worst-case reward under bounded ball attacks without extra perturbation samples. \cite{pattanaik2017robust} integrates gradient-based techniques with Deep Q-Networks (DQN) to assess the impact of perturbations on control performance. Robust CRL, developed from robust RL, addresses constraint satisfaction. \cite{liu2022robustness} uses an on-policy approach for robust optimal control with Lagrangian methods. \cite{liu2023towards} proposes robust CRL with benign data and off-policy EM approaches to avoid disturbances in training. \cite{li2024safe} solves the robust optimal control problem from a dual perspective within the Lagrangian framework. These approaches offer different perspectives on the robust constrained optimal control problem.

\vspace{0.2cm}
Industrial needs are evident, but robust CRL methods for the COFC problem under observational noise remain unaddressed. Hence, the main contributions are as follows:
\vspace{0.15cm}

(i) By modeling observational noise in battery SOC and reference speed, this work extends the COFC problem to the COFC-ON formulation. The formulation explains the noise phenomena frequently encountered in dynamometer testing and PEC design, linking a robust control framework with empirical characteristics.

\vspace{0.08cm}

(ii) A generalized and structured training framework is developed to handle noise in SOC and reference speed. Large-scale simulations on the Prius THS model show that the proposed method achieves consistently low fuel consumption and maintains SOC balance across a range of noise levels, highlighting its resilience to observational disturbances.

\vspace{0.08cm}

(iii) The proposed framework generalizes to testing under custom-designed disturbances beyond observational noise, supporting broader robustness evaluations. It also provides a foundation for extending CRL-based energy control to diverse automotive tasks, including lane keeping, lane changing, and other functions where robustness remains essential.

\vspace{0.15cm}

The remainder of this paper is organized as follows. Section~II provides a brief review of the COFC problem and introduces the COFC-ON formulation. Section~III details the proposed solution approach. Section~IV presents simulation results under uniform observational noise. Section~V concludes the work.

\section{Problem Formulation}

This section provides a brief overview of the COFC problem and develops the COFC-ON formulation. The basic expressions of the COFC-ON problem can also be referenced from~\cite{yancofc} for consistency.

\subsection{Vehicle Dynamics}

We model vehicle dynamics as a discrete nonlinear dynamic system:
\begin{align}
	s_{t+1} &= F(s_{t}, a_{t}) \label{d1} \\
	y_{t} &= Z(s_{t}, a_{t}) \label{d2}
\end{align}
where \( s_{t} \), \( a_{t} \), and \( y_{t} \) are the state, action, and observation vectors at time \( t \), respectively, and \( F \) and \( Z \) are the transition and observation functions. The transition observation is represented as the tuple \( (s_{t}, a_{t}, s_{t+1}, r_{t}, c_{t}) \), where \( r_{t} \) and \( c_{t} \) are the reward and cost, respectively. We denote \( s_{t} \) by \( s \) and \( s_{t+1} \) by \( s' \) for brevity. We also define \( f_t = f(s_t, a_t) \), where \( f \in \{r, c\} \) to represent the reward or cost function. The value function is \( V_f^\pi(\mu_0) = \mathbb{E}_{\tau \sim \pi, s_0 \sim \mu_0} \left[\sum_{t=0}^{\infty} \gamma^t f_t \right] \), which represents the expected discounted return under policy \( \pi \) and initial state distribution \( \mu_0 \). The value function for state \( s \) is \( V_f^\pi(s) = \mathbb{E}_{\tau \sim \pi, s_0=s} \left[\sum_{t=0}^{\infty} \gamma^t f_t \right] \).

In this work, the reward $r_t$ is defined as the negative fuel consumption (in \si{g}) incurred during the transition $(s_t, a_t, s_{t+1})$, with the vehicle dynamics constituting the environment.

\subsection{COFC Problem}

\subsubsection{Description}  
The COFC problem aims to find a policy that maximizes the reward while keeping the cost below a threshold \(\kappa\):  
\begin{align}  
\pi^* = \arg \max _\pi V_r^\pi\left(\mu_0\right), \quad \text{subject to} \quad V_c^\pi\left(\mu_0\right) \leq \kappa.  
\end{align}  

In our work, the COFC problem seeks to minimize the total fuel consumption of an HEV while keeping the battery SOC within the allowed range.

\subsubsection{Constraint} 
\begin{figure}[htbp]
 	\centering
 	\includegraphics[width=3.2 in]{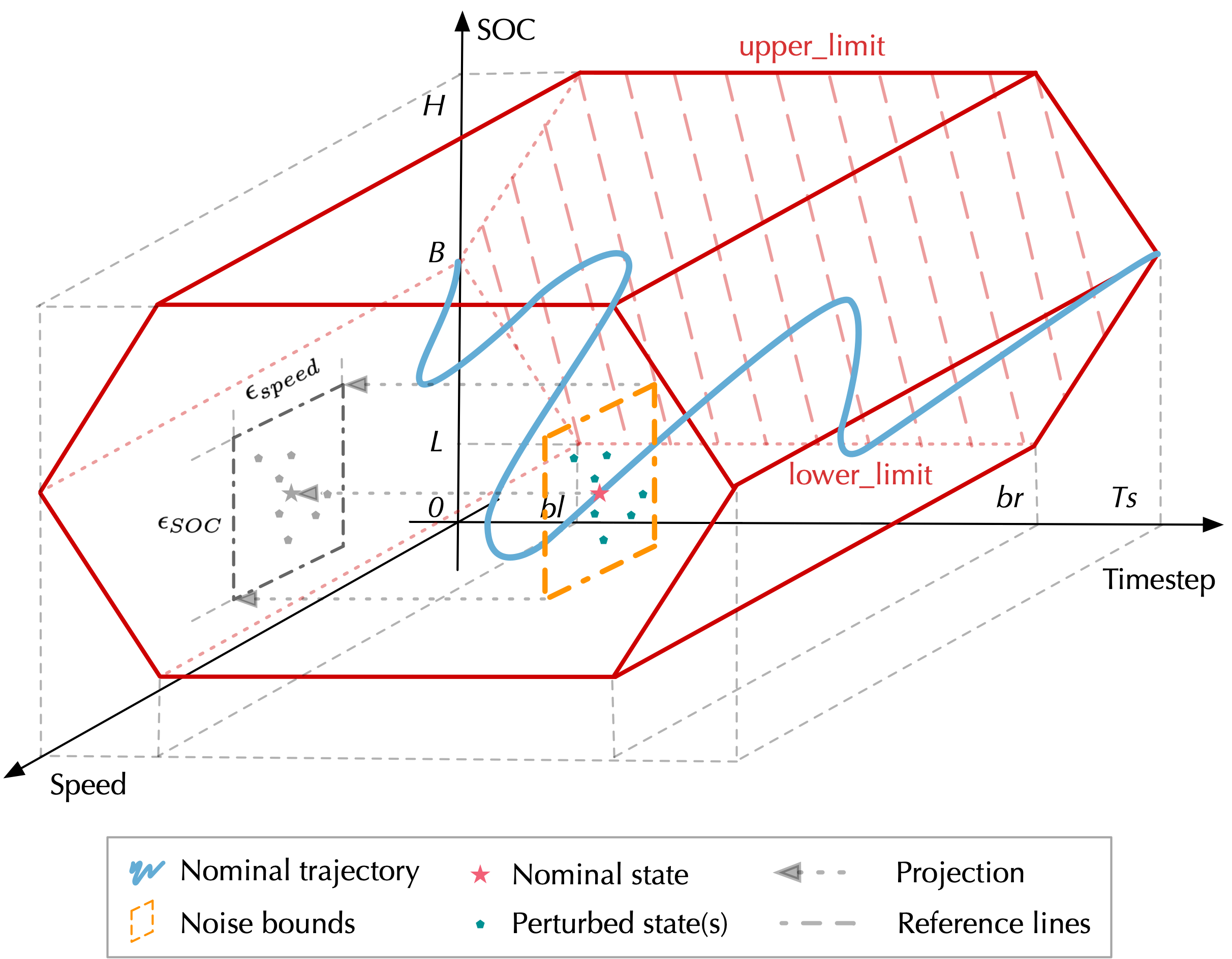}
        \vskip 0.25 cm
 	\caption{Battery SOC constraint region (red polyhedron) with added speed dimension, extended from Fig. 1 of our seminal work~\cite{yancofc}. }
 	\label{mm4}
 \end{figure}

The SOC-allowed range is shown in Fig.~\ref{mm4}, with the upper and lower limits defined by $(\ref{up})$ and $(\ref{low})$, respectively. The values \( H \), \( L \), and \( B \) represent the highest, lowest, and balance point of the allowable SOC range. The vehicle dynamics starts at time step 0 and ends at time step \( Ts \), with critical steps \( bl \) and \( br \) marking changes in trend.

 \begin{align}
 	\text{upper\_limit}=\left\{
 	\begin{array}{rcl}
 		\frac{(H-B)} {bl}  \times t +B ,&& {t      \leq   bl}\\ \\
 		\frac{(H-B)} {br-Ts}  \times (t-Ts) + B,&&{t     >  br} \\ \\
 		H,&&{\text{others}}.
 	\end{array} \right. 
 	\label{up}
 \end{align}
 
 \begin{align}
 	\text{lower\_limit}=\left\{
 	\begin{array}{rcl}
 		\frac{(L-B)} {bl}  \times t  + B,&&{t      \leq    bl}\\ \\
 		\frac{(L-B)} {br-Ts}  \times (t-Ts) + B,&&{t      >     br}\\ \\
 		L, & &{\text{others}}.
 	\end{array} \right. 
 	\label{low}
 \end{align}

Then we set the cost as 
 \begin{align}
 	\text { cost }=\max (\text {SOC}- \text {upper\_limit}, 0) \notag\\
 	+\max (\text {lower\_limit}- \text {SOC}, 0).
 \end{align}
 
We expect the cost to tend toward 0, meaning the SOC dynamics will be within the SOC-allowed range.

\subsection{Description of Observational Noise}
\label{sec:noise}

We introduce a deterministic observational noise function \( \nu(s) \) that distorts the agent's input without altering the environment or nominal trajectory. 
The distorted observation \( \tilde{s} \) lies within an \( \ell_p \)-ball \( B_p^{\epsilon}(s) \) centered at the original observation \( s \), satisfying
\begin{equation}
    \|\tilde{s} - s\|_p \leq \epsilon.
\end{equation}

In practical vehicle applications, although SOC prediction and estimation have advanced, most production vehicles still use simple and low-cost methods. 
SOC is usually estimated from current and voltage measurements by Coulomb counting with OCV correction. 
This approach is robust and easy to implement, whereas advanced estimators such as Kalman filters require high-precision sensors and greater computational power, making them less suitable for large-scale production. 
This reliance on standard sensors highlights the importance of handling measurement noise and uncertainty~\cite{marques2023overview}. 
Similarly, predicted speed profiles are typically provided by map suppliers, but OEMs rarely know the algorithms used to generate them. 
As a result, the deviation between predicted and actual speed remains uncertain and can affect fuel consumption under real driving conditions~\cite{eu2020vectoreport}. 
Such uncertainties motivate the introduction of a controlled observational noise model to bridge perception and the true system state. 

When selecting a noise model, we consider two main factors:
\begin{itemize}
    \item[i)] \textbf{Practicality:} the model should capture essential uncertainties with minimal computational overhead.
    \item[ii)] \textbf{Boundedness:} sensor and actuator errors in vehicles are typically limited within known ranges.
\end{itemize}
Accordingly, a uniform distribution is adopted to represent the engineering noise~\cite{zhang2020robust, mkadry2017towards}:
\begin{equation}
    \|\tilde{s} - s\|_p \sim \mathcal{U}(-\epsilon, +\epsilon).
\end{equation}
Here, $\epsilon$ denotes the noise amplitude, defining a symmetric range centered at zero. 
While both uniform and Gaussian distributions are common in uncertainty modeling, the uniform one better reflects the bounded nature of real sensors. 
Other distributions, including Gaussian or data-driven ones, can also be applied when dynamometer data are unavailable or higher PEC accuracy is required.

\subsection{COFC-ON Problem}
\label{sec:cofcon}

Under observational noise, the trajectories become \(\tilde{\tau} = \{s_0, \tilde{a}_0, \tilde{s}_1, \tilde{a}_1, \dots\}\), 
where actions \(\tilde{a}_t\) are sampled from \(\pi(a \mid \nu(s_t))\). 
The COFC problem is extended to the COFC-ON problem, formulated as:
\begin{equation}
\begin{array}{l}
\hspace{0.5 cm} \pi^* = \arg \max_\pi V_r^{\pi \circ \nu} \left( \mu_0 \right) \\
[0.3cm] 
\text{subject to} \quad V_c^{\pi \circ \nu} \left( \mu_0 \right) \leq \kappa, ~\forall \nu.
\end{array}
\label{op}
\end{equation}

The reward and cost functions are defined as in the COFC problem. 
To solve the COFC-ON problem, we adopt a Lagrangian formulation:
\begin{align}
\left(\pi^{*}, \lambda^{*}\right)=\min _{\lambda \geq 0} \max _{\pi \in \Pi} 
V_{r}^{\pi \circ \nu}\left(\mu_{0}\right)
-\lambda\left(V_{c}^{\pi \circ \nu}\left(\mu_{0}\right)-\kappa\right).
\label{lr}
\end{align}

We follow the Lagrangian-based PPO approach proposed in~\cite{stooke2020responsive}, 
which alternates between maximizing over \(\pi\) and minimizing over \(\lambda\), 
leading to a convergent pair \((\pi_{\text{final}}, \lambda_{\text{final}})\) 
that almost surely\footnotemark[1] converges~\cite{tessler2018reward}. 
This approach is used in our implementation to solve the COFC-ON problem.

\footnotetext[1]{“Almost surely” means the event occurs with probability 1, except on a set of outcomes with probability 0.}

\section{Training Framework}

This section introduces the overall training procedure for solving the COFC-ON problem under observational noise. We first present the general framework that alternates between policy learning and noise adjustment, followed by a specific algorithmic realization based on a robust Lagrangian-based PPO algorithm.

\subsection{Basic Training Procedure}

To solve the COFC-ON problem, we train a CRL agent in an environment with observational noise. As illustrated in Algorithm~\ref{meta}, the agent learns a policy that maximizes reward while satisfying cost constraints, based on trajectories distorted by a noise function. During training, the policy is updated using noisy samples, while the noise scheduler gradually expands the perturbation range~$\epsilon$ to avoid early conservatism and encourage progressive adaptation. This co-adaptive process enhances the robustness and stability of the learned policy.

\begin{algorithm}[htbp]
	\caption{Basic training procedure} 
 	\hspace*{0.02in}{\bf Input:} CRL agent, noise scheduler\\
	\hspace*{0.02in}{\bf Output:} robust policy $\pi$\\
	1:\hspace*{0.02in} Initialize policy $\pi \in \Pi$ and noise function $\nu: \mathcal{S} \rightarrow \mathcal{S}$\\
    2:\hspace*{0.02in} Sample noisy trajectories $\tilde{\tau} = \{s_0, \tilde{a}_0, \ldots\}_T$, with\\ \hspace*{0.15in} $\tilde{a}_t \sim \pi(a \mid \nu(s_t))$\\
 	3:\hspace*{0.02in} {\bf for} $n = 1$ to $N_{\text{train}}$ {\bf do}\\
	4:\hspace*{0.02in}\quad Update policy: $\pi \leftarrow \text{CRL agent}(\tilde{\tau}, \Pi)$\\
	5:\hspace*{0.02in}\quad Adjust noise: $\nu \leftarrow \text{scheduler}(\tilde{\tau}, \pi, n)$\\
	6:\hspace*{0.02in} {\bf end for}
	\label{meta}
\end{algorithm}

Details on how the perturbation range $\epsilon$ evolves are given in the supplementary material or~\cite{githubYanshuchangCRL_COFC_OP}.

\begin{figure*}[htbp]
    \centering
    \hspace{-1 cm}
    \includegraphics[width=14cm]{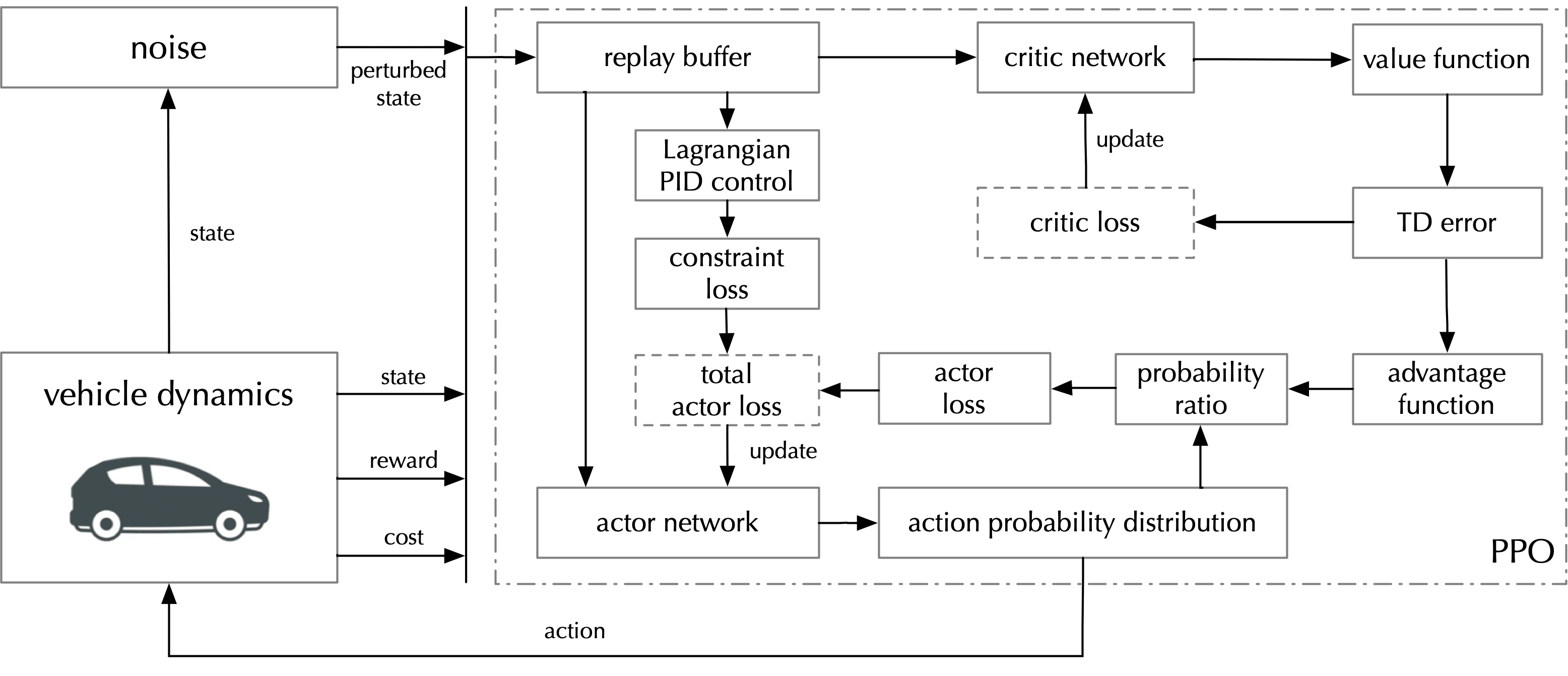}
    \vskip 0.25 cm
    \caption{\small Training procedure: Algorithm~2 as a specific implementation of Algorithm~1 using Lagrangian-based PPO algorithm.}
    \label{fig:training}
\end{figure*}

\subsection{Specific Solving Algorithm}

We adopt the robust PPO-Lagrangian algorithm (Algorithm~\ref{roppo}) to train policies under observational noise. This is a specific instantiation of the general training process in Algorithm~\ref{meta}, designed to handle distorted observations by combining PPO updates with a Lagrangian-based constraint formulation. As shown in Algorithm~\ref{roppo}, the policy networks are trained using trajectories perturbed by a noise function.

\begin{algorithm}[htbp]
	\caption{Robust PPO-Lagrangian Algorithm} 
	\hspace*{0.02in}{\bf Input:}  $T$,  $M$,  $\ell_{\text {ppol}}\left(s, \pi_{\theta}, r, c\right)$,   $\nu(s)$,  $\theta$,  $\phi_{r}$, $\phi_{c}$,  $\phi_{r}^{\prime}$ and  $\phi_{c}^{\prime}$\\
	\hspace*{0.02in}{\bf Output:} policy  $\pi_{\theta}$\\
	1:\hspace*{0.02in} Initialize policy parameters and critic parameters\\
	2:\hspace*{0.02in} {\bf for} each training iteration {\bf do}\\
	3:\hspace*{0.02in}\quad Roll out $T$ trajectories by  $\pi_{\theta} \circ \nu$ from the environment \\ 
    \hspace*{0.23in}$\left\{\left(\nu(s), \nu(a), \nu\left(s^{\prime}\right), r, c\right)\right\}_{N}$\\
	4:\hspace*{0.02in}\quad $\triangleright$ \textit{Update CRL agent}\\
	5:\hspace*{0.02in}\quad  for optimization steps  $m=1, \ldots, M$ {\bf do} \\
	6:\hspace*{0.02in}\quad\quad Compute PPO-Lagrangian loss $\ell_{\text {ppol}}\left(\tilde{s}, \pi_{\theta}, r, c\right)$ \\
	7:\hspace*{0.02in}\quad\quad Update actor $\theta \leftarrow \theta-\alpha \nabla_{\theta} \ell_{ppol}$\\
	8:\hspace*{0.02in}\quad end for\\
	9:\hspace*{0.02in}\quad Update value function based on samples $ \left\{\left(s, a, s^{\prime}, r, c\right)\right\}_{N}$\\
	10:\hspace*{-0.02in}\quad $\triangleright$ \textit{Update noise scheduler}\\
	11:\hspace*{-0.02in}\quad Polyak average the target networks \\
	12:\hspace*{-0.023in}\quad Increase the noise range until it reaches maximum $\epsilon$\\
	13:\hspace*{0.01in} {\bf end for} 
	\label{roppo}
\end{algorithm}

After initialization, each training iteration proceeds as follows. In Step 3, the agent interacts with the environment using the current noisy policy \( \pi_\theta \circ \nu \), collecting distorted trajectories \( \{(\nu(s), \nu(a), \nu(s'), r, c)\}_N \). In Steps 4 to 8, these samples are used to compute the PPO-Lagrangian loss, which balances reward and cost through Lagrange multipliers, and the actor is updated accordingly over multiple optimization steps. In Step 9, the value functions are trained using the original, noise-free transitions \( (s, a, s', r, c) \), ensuring stable estimation of returns and constraint signals. In Steps 10 to 12, the noise range \( \epsilon \) is gradually increased by the scheduler, and the target networks are updated using Polyak averaging to smooth parameter changes across training steps. This procedure enables policy learning under varying noise conditions.

\section{Case Study}

\subsection{Simulation Setup}

\subsubsection{Vehicle Dynamics}

\begin{figure}[htbp]
\captionsetup{justification=raggedright,singlelinecheck=false}
	\centering
	\includegraphics[width=2.8 in]{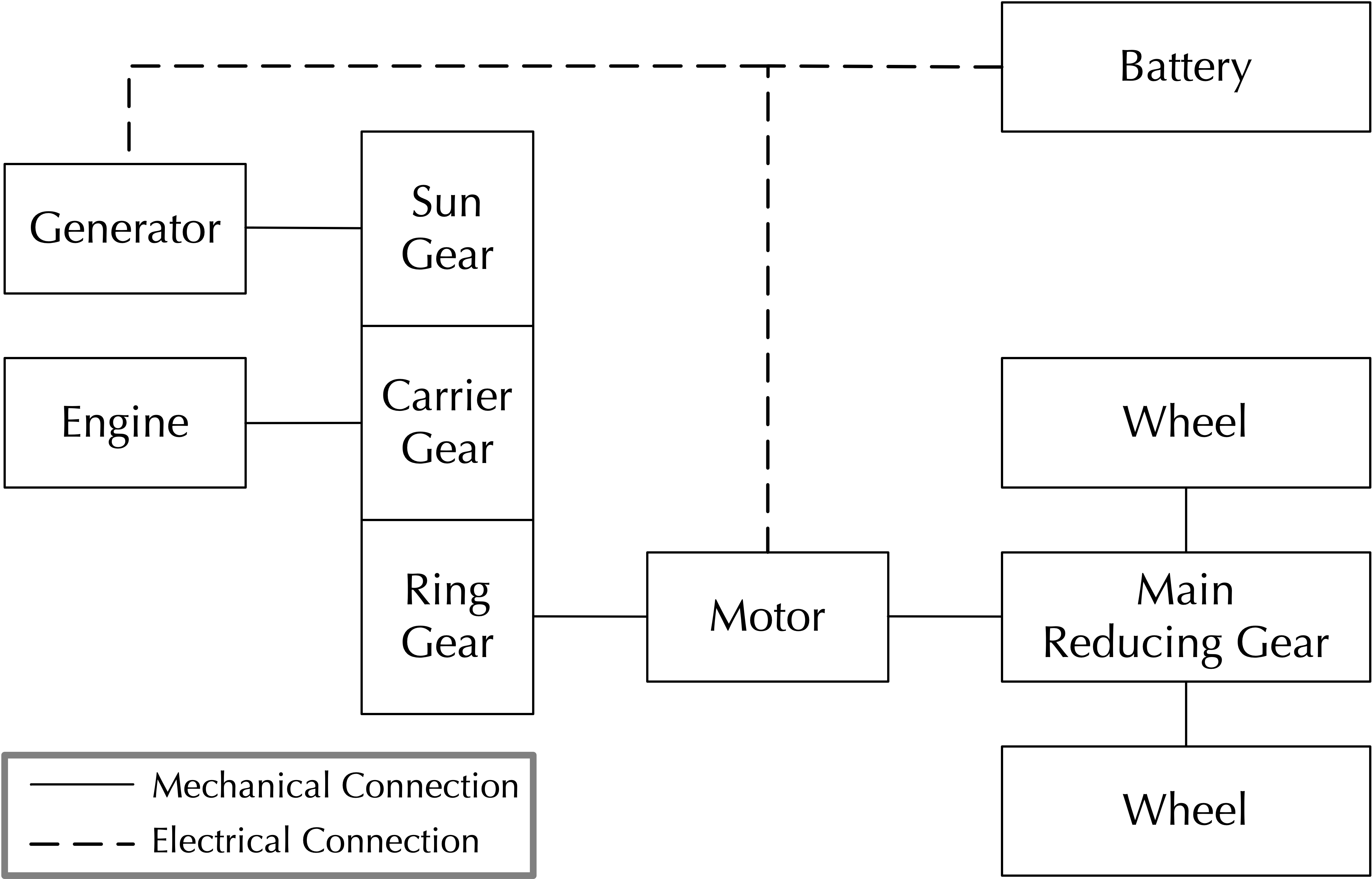}
        \vskip 0.25 cm
	\caption{The Prius THS structure.}
	\label{prius1}
\end{figure}

We use the Prius THS system from~\cite{lian2020rule, liu2023safe, yancofc}, as shown in Fig.~\ref{prius1}, with key parameters from~\cite{lian2020rule}. The model operates in electric and hybrid modes, with observations $\{\text{SOC, velocity, acceleration}\}$ and actions $\{\text{engine power}\}$. Perturbations are applied to SOC and velocity within a 1-norm ball ($\ell_1$-ball), forming a rectangular region in the SOC-velocity space, which is illustrated in Fig.~\ref{mm4}.

The power demand at the wheel during a specified driving cycle or trip is calculated using the longitudinal force balance equation, as shown in Eq.~(\ref{eq1}) below,
\begin{equation}
\hspace{-0.5cm}
\begin{aligned}
P_{\text{wheel}}(t) = v_{car}(t)( \frac{1}{2} C_d A \rho v_{car}(t)^2 + C_{\text{rr}} m_{car} g_{grav} )\\
\left. + m_{car} g_{grav} \sin(\theta_{road}) + m_{car} a_{acc}(t)\right).
\end{aligned}
\label{eq1}
\end{equation}
where $P_{\text{wheel}}(t)$ is the power demand at the wheel [W], $v_{\text{car}}(t)$ is the vehicle speed [m/s], $C_d$ is the drag coefficient, $A$ is the vehicle frontal area [m$^2$], $\rho$ is the air density [kg/m$^3$], $C_{\text{rr}}$ is the rolling resistance coefficient, $m_{\text{car}}$ is the vehicle mass [kg], $g_{\text{grav}}$ is gravitational acceleration [m/s$^2$], $\theta_{\text{road}}$ is the road incline angle [rad], $a_{\text{acc}}(t)$ is the acceleration [m/s$^2$], and $t$ is time [s].

The engine, generator, and motor are modeled using efficiency maps. The battery is represented by an equivalent circuit model, which belongs to the empirical model class~\cite{ramadesigan2012modeling} as follows,
\begin{align} 
P_{\text {batt}}(t) &= V_{\mathrm{oc}}(t)I(t) - R_{0}(t)I^{2}(t) \\
I(t) &= \frac{V_{\mathrm{oc}}(t) - \sqrt{V_{\mathrm{oc}}^{2}(t) - 4R_{0}(t)P_{\text {batt }}(t)}}{2R_{0}(t)} \\
\operatorname{SOC}(t) &= \frac{Q(0) - \int_{0}^{t} I(t) \, dt}{Q_{\text {nom}}}.
\end{align}
where $P_{\text{batt}}(t)$ is the battery power [W], $V_{\mathrm{oc}}(t)$ is the open-circuit voltage [V], $I(t)$ is the current [A], $R_0(t)$ is the internal resistance [$\Omega$], $\operatorname{SOC}(0)$ is the initial state of charge, $Q_{\text{nom}}$ is the nominal battery capacity [Ah], and $\operatorname{SOC}(t)$ is the state of charge. The specific values of these physical and electrical parameters are provided in the supplementary material or at~\cite{githubYanshuchangCRL_COFC_OP}.

\subsubsection{Test Setup}

Fuel consumption is evaluated under both NEDC and WLTC driving cycles. 
WLTC, adopted as China’s official standard in July 2021, features a variable speed profile with an average velocity of 47\,km/h. 
NEDC, averaging 33\,km/h, remains common in the industry due to its closer match with typical urban speeds in China (around 25\,km/h).

Training is performed in Python~3.11~\cite{van1995python} using PyCharm~\cite{JetBrains} on a 2025 MacBook Pro (M4 Pro, 24\,GB RAM) at a controlled temperature of 16\textdegree{}C for thermal consistency. 
NEDC training utilizes all 12 CPU threads, while WLTC is accelerated on Apple silicon GPUs to handle its longer duration and greater speed variability. 
The policy and critic networks adopt a [128, 128] architecture for NEDC and a deeper [256, 256, 256] structure for WLTC, reflecting their different driving dynamics. 
These configurations are determined after repeated testing: smaller networks fail to handle perturbations reliably, whereas larger ones increase training time and memory usage. 
The final settings provide a practical balance between stability and efficiency.

\begin{figure*}[t]
  \centering
  \includegraphics[width=0.9\textwidth]{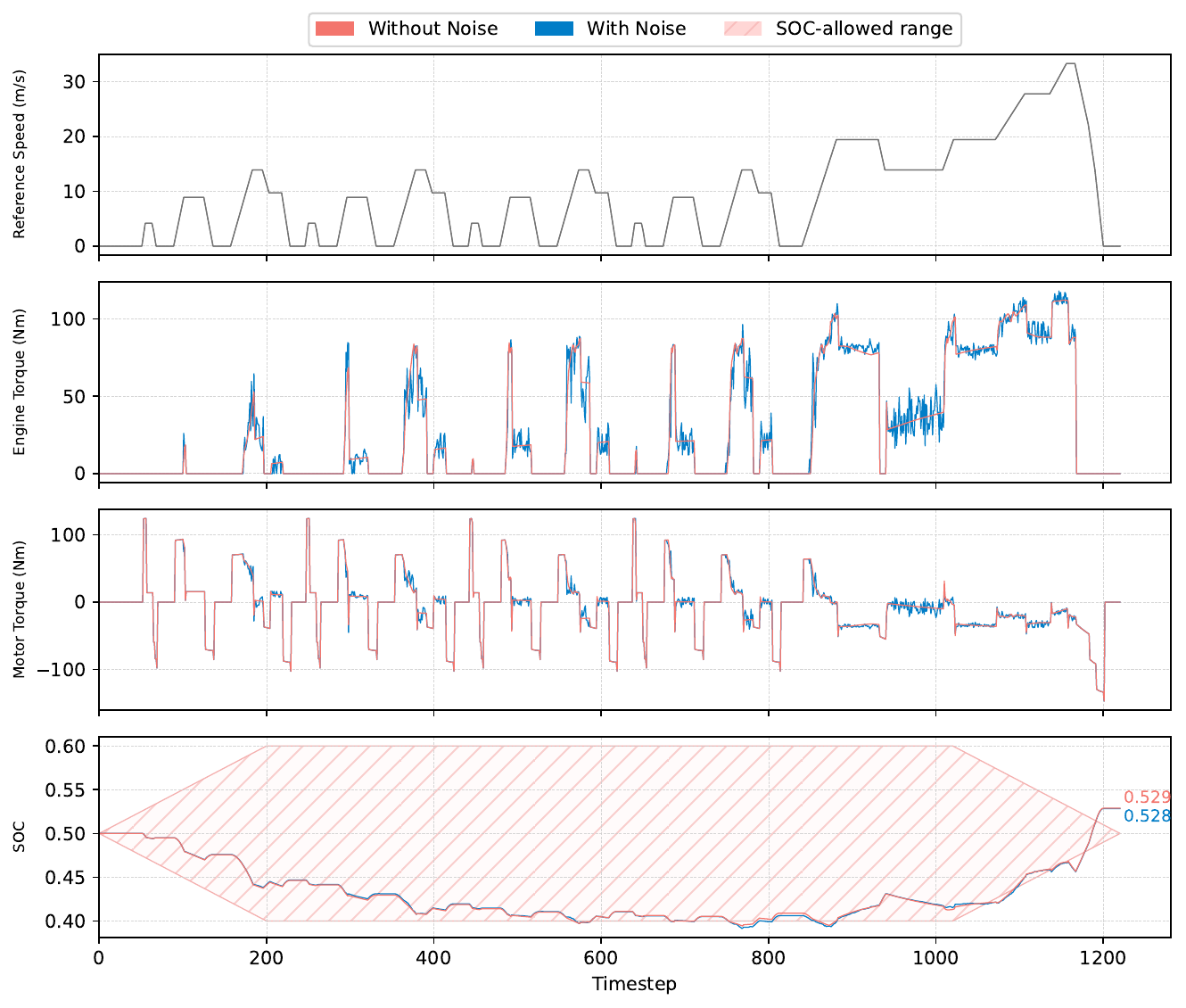}

\caption{Vehicle behavior under the NEDC cycle using the policy trained with the n12 noise configuration (seed 0). Results are shown under both noise-free and noisy conditions, with reference speed, engine torque, motor torque, and battery SOC from top to bottom.}

  \label{dynamics_nedc}
\end{figure*}

\subsubsection{Advances in Testing Accuracy}

SOC observation errors in dynamometer tests mainly arise from sensor precision and environmental variation. 
In production vehicles such as the BYD Qin Plus DM-i, Toyota Prius, and Tesla Model 3, these errors typically fall within 1\%--2\%, consistent with China\textquotesingle s BMS standard GB/T~31467~\cite{GBT31467}. 
Driver speed deviations in HEV testing are also regulated by GB~18352.6–2016~\cite{GB18352.6-2016}, which sets a tolerance of \(\pm 0.56\,\text{m/s}\). 
These standards serve as practical references for setting the perturbation levels in our experiments.

\begin{table}[!ht]
\centering
\renewcommand{\arraystretch}{1.1}
\caption{Simulation Parameters for Multiple Cases}
\label{tab:your_label}
\begin{tabular}{
>{\centering\arraybackslash}p{0.9cm}  
>{\centering\arraybackslash}p{4cm}  
>{\centering\arraybackslash}p{0.2cm}  
>{\centering\arraybackslash}p{2.1cm}  
}
\toprule
Cycle & \(\epsilon_{SOC}\) & \(\epsilon_{speed}\) & \(\nu_{\text{train}}\) / \(\nu_{\text{eval}}\) \\
\midrule
\multirow{4}{*}{\shortstack{NEDC \\ ($\kappa$=1.0)}}
 & \{0.02 (\text{\tiny n3}), 0.01 (\text{\tiny n2}), 0 (\text{\tiny n1})\} & 0 & \scriptsize noise/noise \\
 & \{0.02 (\text{\tiny n6}), 0.01 (\text{\tiny n5}), 0 (\text{\tiny n4})\} & 0.005 & \scriptsize noise/noise \\
 & \{0.02 (\text{\tiny n9}), 0.01 (\text{\tiny n8}), 0 (\text{\tiny n7})\} & 0.01 & \scriptsize noise/noise \\
 & \{\underline{0.02 (\text{\tiny n12})}, 0.01 (\text{\tiny n11}), 0 (\text{\tiny n10})\} & 0.0168 & \scriptsize noise/noise \\
\midrule
\multirow{4}{*}{\shortstack{WLTC \\ ($\kappa$=1.0)}}
 & \{0.02 (\text{\tiny w3}), 0.01 (\text{\tiny w2}), 0 (\text{\tiny w1})\} & 0 & \scriptsize noise/noise \\
 & \{0.02 (\text{\tiny w6}), 0.01 (\text{\tiny w5}), 0 (\text{\tiny w4})\} & 0.005 & \scriptsize noise/noise \\
 & \{0.02 (\text{\tiny w9}), 0.01 (\text{\tiny w8}), 0 (\text{\tiny w7})\} & 0.01 & \scriptsize noise/noise \\
 & \{\underline{0.02 (\text{\tiny w12})}, 0.01 (\text{\tiny w11}), 0 (\text{\tiny w10})\} & 0.0154 & \scriptsize noise/noise \\
\bottomrule
\end{tabular}

\vspace{0.1cm}
\footnotesize
\parbox{\linewidth}{
\textbf{Note:} Items marked as \underline{baseline} indicate the default setting with the largest observation noise, 
which will be further analyzed to understand its impact on vehicle dynamics. 
\(\epsilon_{SOC}\) and \(\epsilon_{speed}\) represent the noise ranges for SOC and speed, bounded within \(\pm 0.02\) and \(\pm 0.56\,\text{m/s}\), respectively. 
The labels n1--n12 and w1--w12 are used for reference and visualization in later plots.
}

\end{table}

\subsection{Simulation Case Setup}

Table~\ref{tab:your_label} summarizes the simulation cases designed to reflect conditions commonly encountered in dynamometer tests. These include various SOC and speed noise levels, aiming to capture realistic sensor errors present in production vehicles. This setup aligns with current engineering practice, where observation noise—particularly in SOC and speed—remains a key challenge.


The noise levels span both typical and upper-bound values. For NEDC and WLTC, the maximum speeds are \(33.33\,\text{m/s}\) and \(36.47\,\text{m/s}\), respectively. The corresponding speed noise ratios (\(\epsilon_{\text{speed}}\)) are set to 0.0168 and 0.0154, keeping deviations within the \(\pm 0.56\,\text{m/s}\) limit defined by national standards. To enhance result reliability, five random seeds (0, 11, 22, 33, and 44) are used. This choice remains within, yet near the upper end of, the typical 3–5 range in RL studies. In total, \(12 \times 5 \times 2 = 120\) simulation cases are run across NEDC and WLTC cycles. The main configurations and results are presented in this paper.

\begin{figure*}[t]
  \centering
  \includegraphics[width=0.9\textwidth]{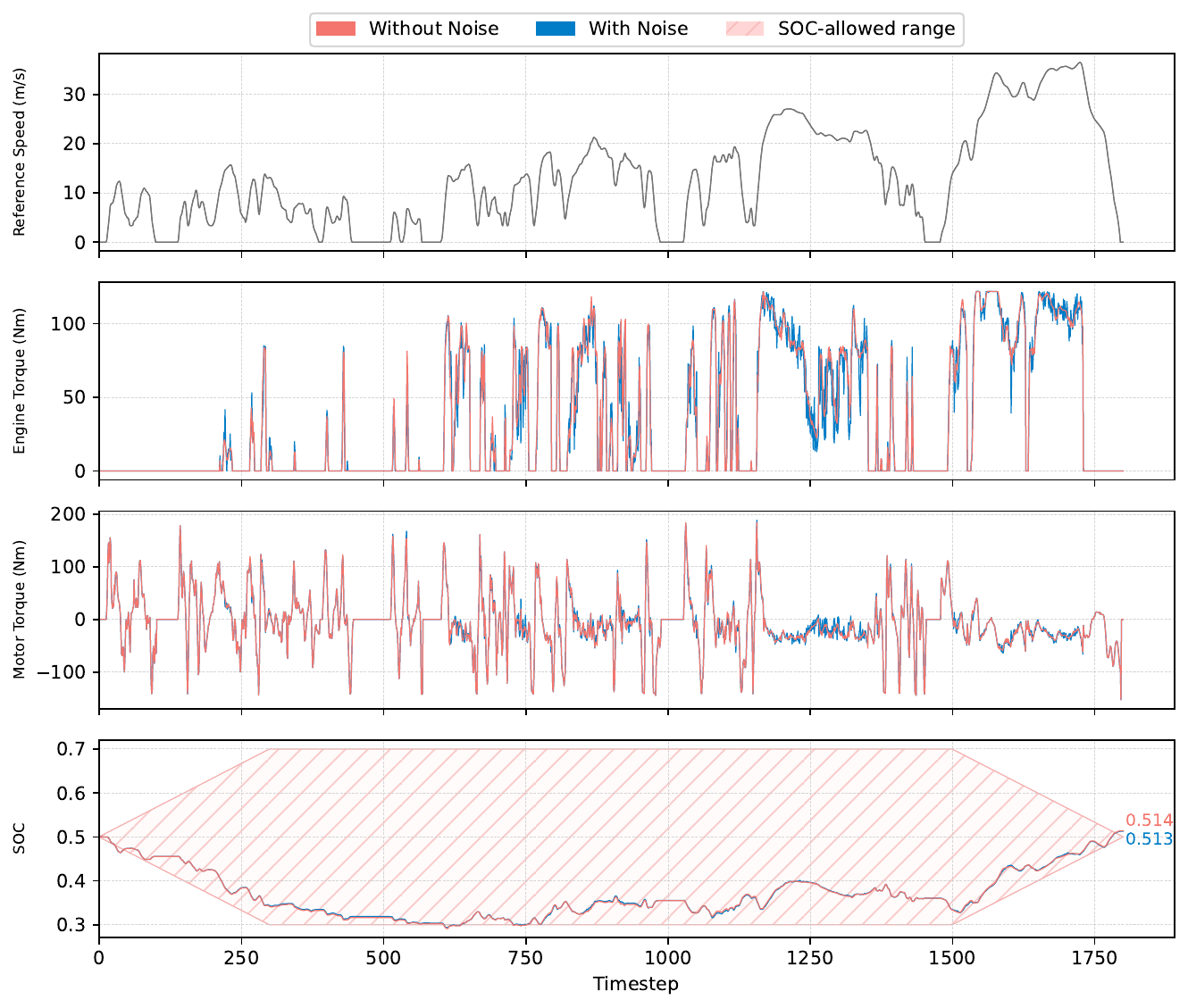}
\caption{Vehicle behavior under the WLTC cycle using the policy trained with the w12 noise configuration (seed 0). Results are shown under both noise-free and noisy conditions, with reference speed, engine torque, motor torque, and battery SOC from top to bottom.}
  \label{dynamics_wltc}
\end{figure*}

\subsection{Results and Analysis}

\subsubsection{Performance Validation of Baseline}

To clarify the explanation and focus our analysis, we selectively present the results of the final policy trained under the highest noise configurations---n12 for NEDC and w12 for WLTC---using seed 0. To assess how observational noise affects execution, we deploy this policy in two environments: one with the same noise level used during training, and one without any noise. \textcolor{mc1} {Figures~\ref{dynamics_nedc} and~\ref{dynamics_wltc}} show the reference speed, engine torque, motor torque, and battery SOC trajectories. The results illustrate how the same learned policy behaves differently depending on the presence or absence of noise.

\textit{(a) Analysis of Vehicle Behavior under NEDC:} In the no-noise case, the initial SOC is also 0.500 and the final SOC reaches 0.529, with a total fuel consumption of 332.645 g and an SOC violation cost of 1.067. In the noise case, the vehicle starts from an initial SOC of 0.500 and ends at 0.528, with a slightly higher fuel consumption of 334.733 g and an SOC violation cost of 1.233. Despite the presence of noise, the increase in fuel consumption is marginal. In both cases, the final SOC remains within the industry-accepted range of $0.50 \pm 0.03$.



For the engine performance, the engine and motor collaborate to meet torque demands throughout the driving cycle. During the suburban phase (850--1167\,s), the engine becomes the dominant source due to higher speeds. In the earlier urban segments (0--780\,s), the motor and engine complement each other to meet low-speed demands. When the speed drops below 4.2\,m/s, the motor takes the lead, while the engine provides auxiliary support. In the no-noise case, engine torque increases smoothly during acceleration events, such as at 172--185\,s, 291--299\,s, 361--377\,s, 485--490\,s, 556--574\,s, 678--686\,s, and 748--768\,s. Under noise, however, the torque trajectory exhibits high-frequency oscillations across the entire cycle, particularly during steady high-speed intervals like 364--391\,s, 560--586\,s, and 941--1009\,s, where the noise-free torque remains stable while the noisy counterpart fluctuates around it. This behavior closely resembles patterns commonly observed in real-world tests under the influence of noise.

For the motor performance, the motor primarily delivers driving torque at low speeds and regenerative torque during deceleration. In the first four urban cycles (0--780\,s), when the reference speed is below 4.2\,m/s, it serves as the main torque provider. In the no-noise case, the torque profile is more distinct—positive during acceleration and negative during braking—clearly reflecting energy recovery during deceleration phases. Under noise, torque fluctuations become more pronounced during these braking intervals (171--183\,s, 361--379\,s, 556--571\,s, 751--768\,s, and 850--881\,s), as well as during steady-speed segments of the suburban cycle (942--1021\,s), with frequent switching between positive and negative values. This suggests stronger motor intervention to compensate for variations in engine output.

For the battery performance, the SOC trajectories in both the no-noise and noise cases appear smooth at the current plotting scale, with no significant oscillations observed in the noise case. Minor differences occur during 290--375\,s and 754--856\,s, but the overall patterns remain consistent: SOC first decreases and then increases, staying within the predefined allowable range. Specifically, the minimum SOC in the no-noise case is 0.394 at 769\,s, while in the noise case it drops slightly lower to 0.391 at 768\,s. This suggests that the controller effectively maintains SOC balance and satisfies constraints under observational noise.


\begin{figure*}[t]
  \centering
  \includegraphics[width=\textwidth]{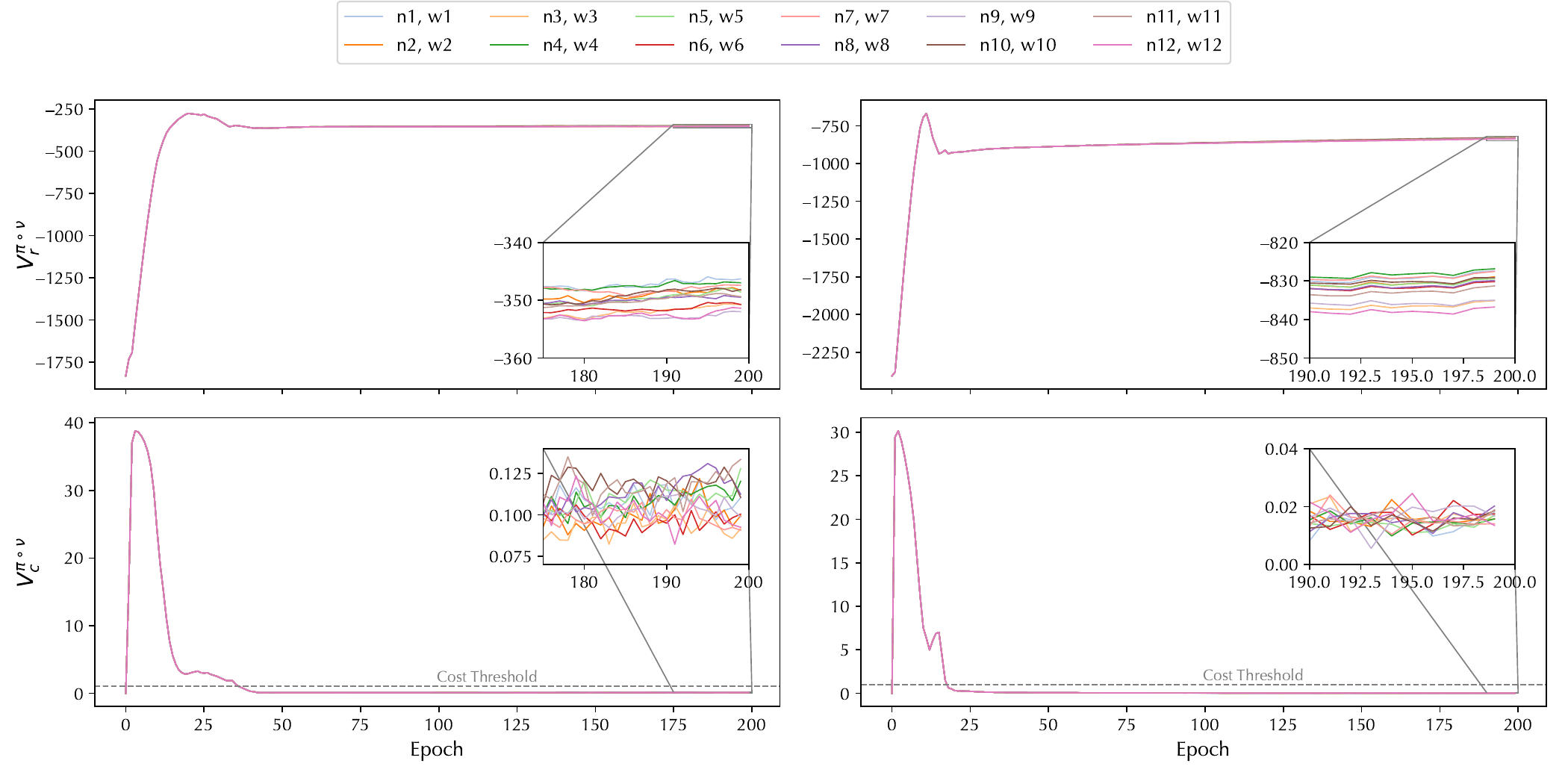}

\caption{Value functions $V_{r}^{\pi \circ \nu}$ and $V_{c}^{\pi \circ \nu}$ across noise configurations n1–n12 and w1–w12 under NEDC (left column) and WLTC (right column). Insets highlight differences in training trajectories under various perturbation scales. For each configuration, the plotted trajectories represent the average over five representative random seeds.}

  \label{wltc}
\end{figure*}

\textit{(b) Analysis of Vehicle Behavior under WLTC:} In the no-noise case, the vehicle starts from an initial SOC of 0.500 and ends at 0.514, with a total fuel consumption of 769.090\,g and an SOC violation cost of 0.357. In the noise case, the SOC also begins at 0.500 and ends at 0.513, with a slightly higher fuel consumption of 770.871\,g and an SOC violation cost of 0.291. Despite the presence of noise, the difference in fuel consumption is marginal. In both cases, the final SOC remains within the industry-accepted range of $0.5 \pm 0.03$.



For engine performance, the torque variation closely follows the reference speed profile. During high-speed phases—1168\,s to 1350\,s and 1537\,s to 1732\,s—where the speed exceeds 20\,m/s, the engine provides the dominant propulsion, with torque mostly above 80\,Nm. Under noise, the torque oscillates around the baseline, with fluctuations mostly within $\pm$19\,Nm. At the beginning of the cycle (0–500\,s), the engine only occasionally engages, as the propulsion demand is primarily met by the motor.

For motor performance, during the high-speed segments (835–950\,s, 1168–1363\,s, and 1506–1744\,s), the motor torque in both cases remains close to zero and is mostly negative. In these periods, the engine supplies most of the driving force, while the motor primarily contributes through regenerative braking. Under noise, the motor torque exhibits clear fluctuations around the no-noise baseline. During other intervals, the motor and engine jointly provide torque, with motor torque ranging approximately from --150\,Nm to 150\,Nm. In the earlier segments (0--600\,s), the motor serves as the primary torque source.

For battery performance, the SOC exhibits a consistent downward-then-upward trend. In the no-noise case, it decreases from 0.500 to a minimum of 0.291 at 612\,s, then recovers to 0.514 by the end of the cycle. In the noise case, the minimum value is slightly higher at 0.293, also occurring at 612\,s, with a final value of 0.513. Localized differences between the two curves emerge around 280\,s, 1088\,s, and 1585\,s, where noise introduces subtle yet discernible deviations. Nevertheless, neither case shows high-frequency oscillations. The controller effectively maintains the SOC within the desired bounds and successfully drives the trajectory toward the target level despite the presence of observation noise


The control profiles analyzed in (a) and (b) represent the theoretically derived engine and motor trajectories under the NEDC and WLTC cycles. Although optimal in theory, these trajectories still need to be smoothed---such as by applying a low-pass filter---before they can be used in real vehicles. This kind of smoothing is also common in rule-based control strategies. We observe that under observational noise, the engine and motor torque trajectories fluctuate around their noise-free versions. This is a natural and interesting effect that shows how noise can disrupt smooth control behavior. In contrast, the battery SOC trajectories exhibit overall stability, with gradual trends and occasional bends that reflect the controller’s effort to regulate SOC under noisy observations.

\begin{figure*}[htbp]
    \centering

    \begin{minipage}[t]{0.48\textwidth}
        \centering
        \includegraphics[width=\textwidth]{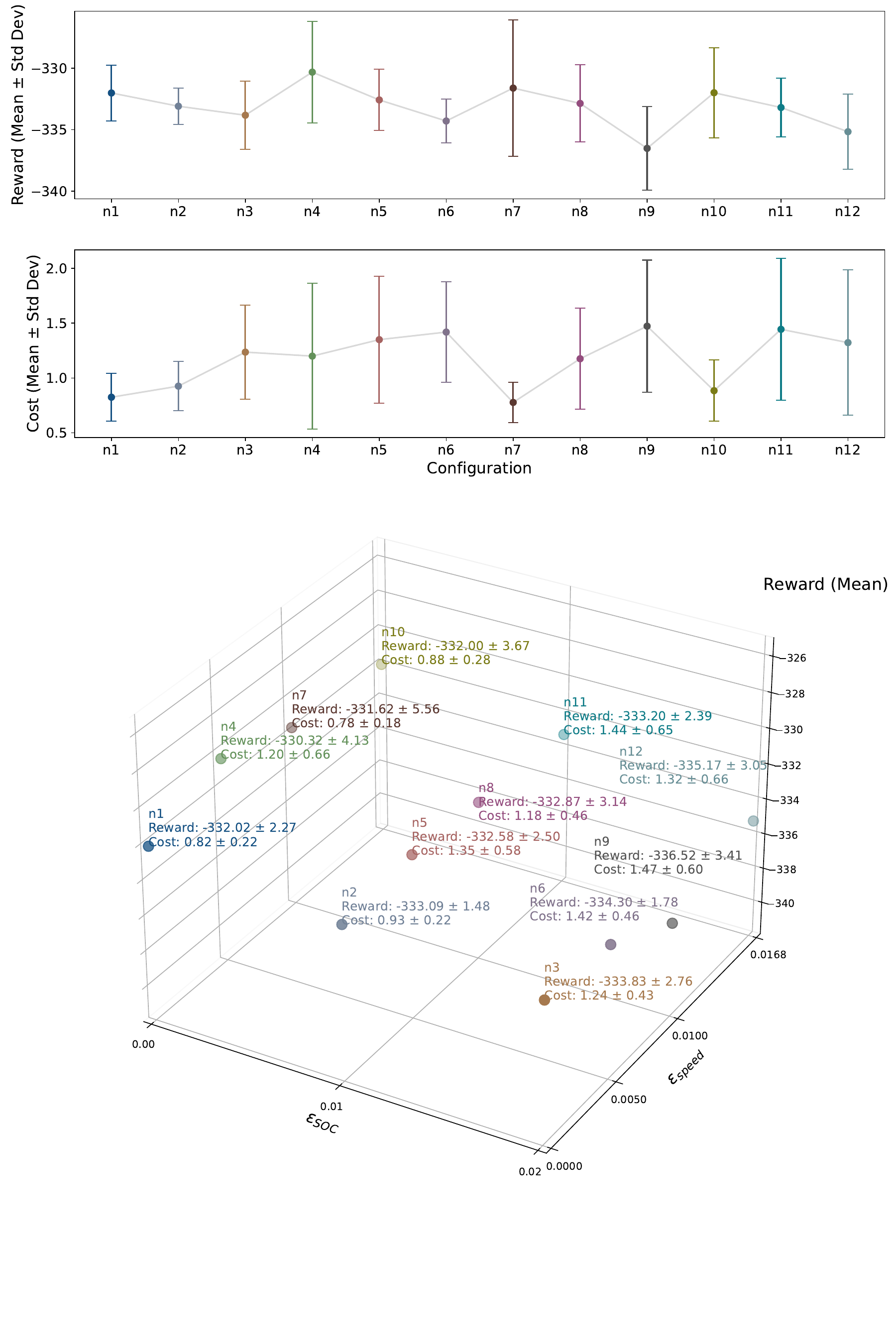}
        \vspace{1 ex}
        \vskip -1.5 cm
        \label{fig:nedc}
    \end{minipage}
    \hfill
    \begin{minipage}[t]{0.48\textwidth}
        \centering
        \includegraphics[width=\textwidth]{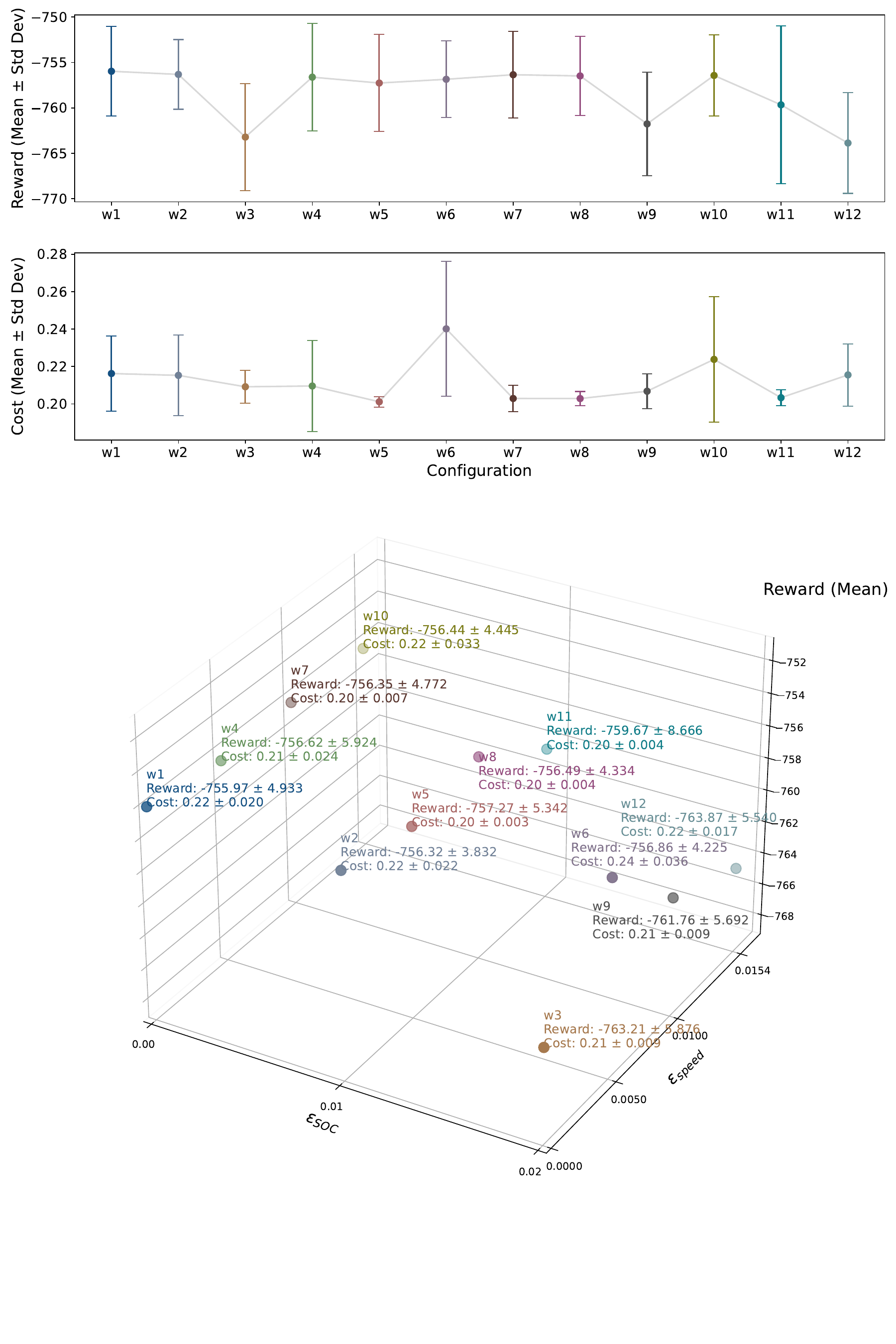}
        \vspace{1 ex}
        \vskip -1.5 cm
        \label{fig:wltc}
    \end{minipage}
    \vskip -1.0 cm
\caption{Simulation results under two representative driving cycles: NEDC (left column) and WLTC (right column). Each column includes three vertically arranged plots: the mean reward with standard deviation (top), the mean cost with standard deviation (middle), and a 3D scatter view of reward versus SOC and speed noise perturbations (bottom).}

    \label{fig:comparison_nedc_wltc}
\end{figure*} 

\subsubsection{Noise-Scale Influence on Fuel Consumption}

We examine how SOC and speed noise affect policy performance, focusing on training stability and fuel consumption under different noise scales.

\textit{(a) Training Performance under Observational Noise:}
Each configuration (n1–n12 for NEDC and w1–w12 for WLTC) was independently trained with five random seeds. At each training epoch, the average return and cost across the five runs were computed to represent typical performance. The results are shown in Figure~\ref{wltc}.

The cumulative reward  $V_{r}^{\pi \circ \nu}$ rises quickly within the first 25 epochs and stabilizes by around epoch 50, with final values between $-360$ and $-340$. The constraint value $V_{c}^{\pi \circ \nu}$ drops below the threshold early and remains under 0.15 for most configurations. Even under the strongest perturbations (n12 and w12), the training process remains stable. Across all configurations, the overall training curves are nearly identical. However, local zoom-in reveals differences depending on the perturbation. For both NEDC and WLTC, the gap between the highest and lowest reward trajectories within the zoomed-in region is within 10. For cost, the variation in the same region is about 0.05 for NEDC and 0.03 for WLTC.

Since  $V_{r}^{\pi \circ \nu}$  and $V_{c}^{\pi \circ \nu}$ reflect the estimated discounted reward and cost, these results confirm that training remains stable and effective across all tested cases.  In analysis~(b), we assess fuel consumption based on the trained policies.

\textit{(b) Fuel Consumption across Perturbation Scales:} To evaluate fuel economy, we apply the trained policy—recovered under its corresponding training noise level—from the last five epochs of each satisfactory seed to the vehicle model, aiming for a more stable estimate. We then compute the mean fuel consumption, SOC violation cost, and their standard deviations. Results for both NEDC and WLTC are shown in Figure~\ref{fig:comparison_nedc_wltc}.

Under the NEDC cycle, mean rewards range from $-336.52$ to $-330.32$, with most values clustered around $-333$, indicating stable fuel consumption under nationally regulated noise levels. Strong SOC noise (e.g., n3, n6, n9, n12) results in lower rewards and thus higher fuel use. Configuration n4 achieves the best reward ($-330.32$), though its cost is slightly higher (1.20). As SOC noise increases, fuel consumption rises and reward declines. From n1 to n3, the reward drops from $-332.02$ to $-333.83$, and the cost rises from 0.82 to 1.24. From n4 to n6, the reward decreases from $-330.32$ to $-334.30$, with costs from 1.20 to 1.42. For n7 to n9, the reward falls from $-331.62$ to $-336.52$, and the cost increases from 0.78 to 1.47. From n10 to n12, the reward varies from $-332.00$ to $-335.26$, with costs ranging from 0.88 to 1.44. For a fixed SOC level, speed noise does not show a discernible trend in fuel consumption.

Under the WLTC cycle, the policy becomes more sensitive to noise. The reward spans from $-763.87$ to $-755.97$, with wider dispersion and larger error bars compared to NEDC, reflecting the higher-speed, more dynamic nature of WLTC. From w1 to w3, the reward drops from $-755.97$ to $-763.21$, and the cost ranges from 0.21 to 0.22. From w10 to w12, the reward decreases from $-756.44$ to $-764.14$, with costs ranging from 0.20 to 0.22. For other groups like w4–w6 and w7–w9, the fuel consumption is not too sensitive to the SOC noise increase. Configurations with maximum SOC noise, such as w3 and w12, result in significantly degraded performance, with rewards of $-763.21$ and $-764.14$, and costs of 0.21 and 0.22, respectively.

The 3D reward landscapes show that reward does not always decrease with increasing noise. Some moderate-noise configurations, such as w5, form local minima, indicating non-monotonic trends. Specifically, w5 has a relatively low mean reward of -757.27 with a mean cost of 0.20. Moreover, reward and cost are not strictly coupled---some settings achieve low fuel use with moderate cost. For example, configuration n4 achieves the lowest fuel consumption, with the highest mean reward (-330.32), but its mean cost of 1.20 is neither the highest nor the lowest among all settings, further illustrating this trade-off. This reflects the policy’s ability to adjust its strategy based on the noise level, balancing fuel consumption and constraint satisfaction.

While performance varies across conditions, the policy remains consistently effective—showing stronger tolerance to speed noise under NEDC and better adaptability to SOC noise under WLTC. These differences highlight the interplay between driving cycles and noise. Although fuel consumption differences may seem small, they still demonstrate the value of the proposed approach. Each engineering problem calls for a clear problem description and a method suited to its specific noise characteristics, as robustness is equally important. The stable results under standardized noise levels confirm the reliability of the proposed framework for practical HEV applications.

\section{Conclusion}

This work presents the first formal formulation of the COFC-ON problem and addresses it using a robust constrained reinforcement learning approach that explicitly accounts for observational noise. The results show that moderately sized policy and value networks—with two to three layers per network and up to 256 units per layer—can effectively handle noise in SOC and speed as specified by Chinese national standards. Under both NEDC and WLTC cycles, observational noise leads to differences in fuel consumption, though the extent of impact varies between the two cycles.

By incorporating observational noise into both the problem formulation and the training process, the proposed method improves the robustness of HEV evaluation and control. These insights pave the way for developing control strategies that achieve optimal fuel consumption under practical observation inaccuracies, such as SOC sensing drift and speed reference mismatch.

\ifCLASSOPTIONcaptionsoff
\newpage
\fi

\bibliographystyle{IEEEtran}
\bibliography{journal}
\vspace{11pt}

\vfill

\end{document}